\documentclass[runningheads]{llncs}
\usepackage[T1]{fontenc}
\usepackage{booktabs}
\usepackage{graphicx}
\usepackage{amsfonts}
\usepackage{url}
\usepackage{xcolor}
\usepackage{lipsum}
\usepackage{subfigure}
\usepackage{algorithm}                    
\usepackage[noend]{algpseudocode}         


\newcommand{\algorithmicinput}{\textbf{Input:}}
\newcommand{\algorithmicoutput}{\textbf{Output:}}
\newcommand{\INPUT}{\item[\algorithmicinput]}
\newcommand{\OUTPUT}{\item[\algorithmicoutput]}

\newcommand\blfootnote[1]{%
  \begingroup
  \renewcommand\thefootnote{}\footnote{#1}%
  \addtocounter{footnote}{-1}%
  \endgroup
}

\begin{document}
\title{Scalable Multi-robot Motion Planning via Hierarchical Subproblem Expansion and Workspace Decomposition Refinement}
\titlerunning{CIPHER}
%
\author{Isaac Ngui*\inst{1}\orcidID{0009-0009-3580-0220} \and
Courtney McBeth*\inst{1}\orcidID{0000-0002-6826-0349} \and
James D. Motes\inst{1}\orcidID{0000-0002-9553-7331} \and
Marco Morales\inst{1,2}\orcidID{0000-0003-1824-2350} \and 
Nancy M. Amato\inst{1}\orcidID{0000-0001-5817-5290}}
\authorrunning{I. Ngui et al.}
%
\institute{University of Illinois Urbana-Champaign, Urbana IL 61801, USA\\
\email{\{ingui2, cmcbeth2, jmotes2, moralesa, namato\}@illinois.edu}
\and
Instituto Tecnológico Autónomo de México, Mexico City, Mexico}
\maketitle              
\begin{abstract}
A fundamental challenge in multi-robot motion planning is achieving sufficient coordination to avoid inter-robot conflicts without incurring the large computational expense of searching the joint configuration space of the robot group.
In this work, we present a method for multiple mobile robot motion planning that achieves an improvement in planning time up to an order of magnitude by leveraging the insight that we can use discrete search over a workspace decomposition to provide coordination between robots during planning. While prior work uses workspace topology to inform when coordination between robots is needed and then composes robots into their joint configuration space, we take a step further by iteratively refining our workspace representation to allow our planner to search smaller, decoupled configuration spaces.

\blfootnote{*Equal Contribution}

\end{abstract}
\section{Introduction}
Multi-robot systems have applications ranging from autonomous warehouse management to search and rescue. These applications require careful multi-robot motion planning (MRMP) to avoid collisions with the environment and between robots.
There exists a tradeoff between the amount of coordination provided during planning and planning speed, especially considering large robot groups with kinodynamic constraints.
Searching the joint configuration space of the robot team, the \textit{composite space}, ensures collision avoidance but comes at a significant computational cost due to the size of the space.
Alternatively, decomposing the problem requires additional reasoning to prevent unresolved inter-robot conflicts.
Existing approaches have explored using \textit{guidance}, often provided by a workspace representation, to expedite MRMP. 
Prior works~\cite{mcbeth2023cdrrrt,mcbeth2023hypergraph} use \textit{sampling guidance} to bias sampling toward high-quality areas of the composite space.
Several existing planners~\cite{mcbeth2023hypergraph,wagner2012srrt,solis2024arc} leverage \textit{coordination guidance} where the planner reasons over when robots should be composed into the same configuration space.
Often, this is indicated by a topological skeleton of the workspace, which is an embedded graph where edges represent free areas of the workspace.
While skeletons provide useful guidance to motion planners in environments with narrow passages, they struggle to represent open environments well, leading us to consider a region decomposition~\cite{plaku2010syclop} to make our approach robust to environments with open spaces.
These forms of guidance reduce the computational burden of providing coordination during planning, but still require planning in the composite space of multiple robots either for the entirety of planning or when inter-robot conflicts cannot otherwise be prevented.


In this work, we take a step toward avoiding the computational bottleneck of planning in the composite space by introducing \textit{resolution guidance} which provides coordination by reasoning over the workspace representation at different resolutions. Our proposed MRMP approach, \textit{Coordinated Incremental Planning with Hierarchical Expansion and Refinement (CIPHER)}, adaptively refines the resolution of the workspace representation. By directing the movement of the robots over the workspace representation we provide coordination at a high level and reduce the need to plan in the composite space.

\begin{figure}
    \centering
    \hfill
    \subfigure[Decomposition]{
        \includegraphics[width=0.2\textwidth]{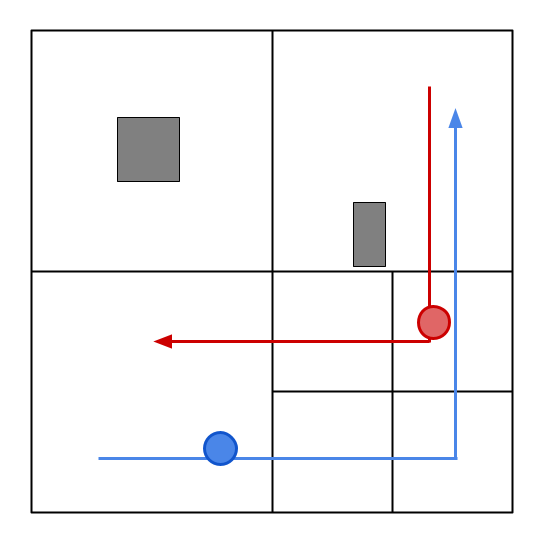}
    }
    \hfill
    \subfigure[Skeleton]{
        \includegraphics[width=0.2\textwidth]{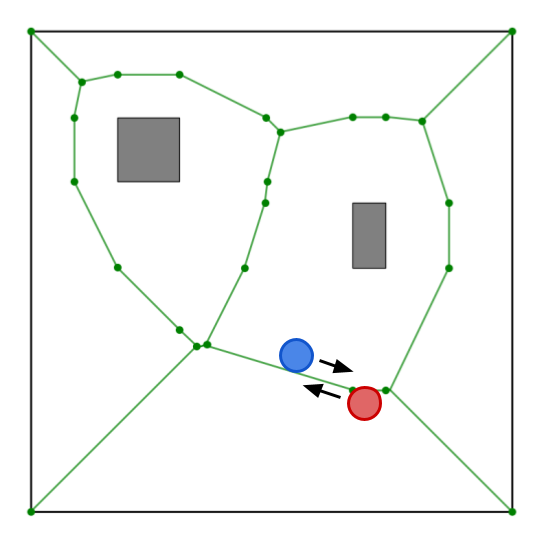}
    }
    \hfill
    \hfill
    \caption{(a) An example of iterative decomposition of the workspace representation. Robots' paths over the region decomposition (colored lines) attempt to avoid coming into proximity with other robots. (b) A skeleton representation for the same environment. Despite the open space, robots are forced into proximity with each other by skeleton sampling guidance, increasing risk of conflicts.}
    \label{fig:reps}
    \vspace{-10pt}
\end{figure}

We additionally recognize that we can refine our region decomposition during planning, leveraging resolution guidance.
CIPHER features a hierarchical search, planning high-level paths for each robot over the workspace representation and then translating those into paths through configuration space.
When multiple robots traverse a region, we attempt to further decompose that region to separate robots onto distinct paths over the workspace representation (Fig.~\ref{fig:reps}).
By doing this, we provide coordination between robots at a high level and reduce the need to compose robots into larger, computationally expensive configuration spaces.
Our method shows an improvement in planning time up to an order of magnitude compared to other state-of-the-art geometric and kinodynamic MRMP approaches.
Our contributions include:
\begin{itemize}
    \item CIPHER, a hybrid multi-robot motion planning algorithm that searches over a region decomposition to coordinate robots at a high level, reducing the need to plan in the composite space.
    \item Resolution guidance, a process by which our method iteratively refines the workspace representation to direct the movement of robots at a finer level.
    \item An extensive experimental validation that evaluates the strengths and weaknesses of our approach compared to other methods.
\end{itemize}

\section{Related Work}
In this section, we discuss related work in the fields of multi-robot motion planning and guided motion planning. We highlight strengths and weaknesses of existing approaches relative to our work.

\subsection{Multi-robot Motion Planning}
Multi-robot motion planning methods can be categorized as \textit{coupled}, \textit{decoupled}, or \textit{hybrid}. Coupled methods~\cite{sl-uppccdpmrs-2002} plan paths by searching the composite space. While they provide high coordination, they lack scalability due to the large size of the composite space. Decoupled methods~\cite{sl-uppccdpmrs-2002,bennewitz2001prioritized} plan for each robot separately, often using a priority ordering. These methods achieve high scalability but do not have the level of coordination required to solve complex problems. Hybrid methods seek to leverage the advantages of both coupled and decoupled planning. These methods leverage different forms of coordination guidance to decide when to search the composite space of multiple robots. We focus our discussion of multi-robot motion planning on offline hybrid planners and kinodynamic planners.

\begin{table}[]
\centering
\begin{tabular}{|l|l|l|l|l|l|}
\hline
                                                              & \begin{tabular}[c]{@{}l@{}}Decoupled \\ Kinodynamic\\ RRT\end{tabular} & \begin{tabular}[c]{@{}l@{}}Coupled \\ Kinodynamic\\ RRT\end{tabular} & K-ARC      & K-WG-DaSH  & \textbf{CIPHER}                                                          \\ \hline
Composition                                                   & decoupled                                                              & coupled                                                              & hybrid     & hybrid     & hybrid                                                                   \\ \hline
Guidance Type                                                 &                                                                        &                                                                      & kinematic  & skeleton   & \textbf{\begin{tabular}[c]{@{}l@{}}cellular\\ decomp\end{tabular}}       \\ \hline
\begin{tabular}[c]{@{}l@{}}Conflict\\ Resolution\end{tabular} & \begin{tabular}[c]{@{}l@{}}prioritized\\ planning\end{tabular}         &                                                                      & subproblem & subproblem & \textbf{\begin{tabular}[c]{@{}l@{}}guidance +\\ subproblem\end{tabular}} \\ \hline
\end{tabular}
\caption{Multi-robot Kinodynamic Planner Characteristics}
\label{tab:kinodynamic-planners}
\vspace{-30pt}
\end{table}

\subsubsection{Hybrid Planners}
Hybrid planners combine coupled and decoupled planning. Several methods do this by searching the individual robots' configuration spaces first and then expanding the search space to provide more coordination  as needed. Subdimensional expansion RRT (sRRT)~\cite{wagner2012srrt} plans RRTs in the individual configuration spaces of each robot and follows those until inter-robot conflicts occur. Then, it expands the planning space to plan for the affected robots in their joint configuration space until they reach their goals. Adaptive Robot Coordination (ARC)~\cite{solis2024arc} similarly plans decoupled paths, but uses local subproblems to resolve inter-robot conflicts when they occur. It attempts to use less expensive planners before resorting to replanning the relevant portion of the paths in the composite space.
Discrete RRT (MRdRRT)~\cite{solovey2016drrt} builds PRMs for each robot and searches the implicit composite space of the robots by building a tensor product roadmap.
The scalability of these approaches shows the benefits of planning in smaller configuration spaces when possible, however, they do not support kinodynamic constraints and reactively account for collisions rather than proactively.

\subsubsection{Kinodynamic Planners}
Kinodynamic planners are required to find plans for robots that adhere to a set of kinodynamic constraints.
Many single-robot planners have extended classical sampling-based algorithms to solve problems with kinodynamic constraints, like Kinodynamic RRT~\cite{hklr-rkmpwmo-02,karaman2010optimal,o-idbrrt-24}, with many preserving the properties of sampling-based algorithms such as probabilistic completeness and asymptotic optimality.
Additionally, many of these algorithms have also been extended to the multi-robot motion planning problem by leveraging multi-agent path finding techniques, such as conflict-based search (CBS)~\cite{SHARON201540}, to efficiently resolve conflicts~\cite{kal-cbsmrmp-22,moldagalieva2024dbcbs,zprcr-cbgcs-25}.
Kinodynamic CBS (K-CBS)~\cite{kal-cbsmrmp-22} presents a straightforward extension of Kinodynamic RRT to the multi-robot motion planning problem, where Kinodynamic RRT is utilized as the low-level planner of CBS. 
Similarly, Discontinuity-bounded CBS (db-CBS)~\cite{moldagalieva2024dbcbs} utilizes the sequences of discontinuous motion primitives sampled using RRT where the discontinuities are bounded by a specified value.
Then, the discontinuous trajectories are fixed using gradient-based trajectory optimization which minimizes the discontinuity between the motion primitives.
This allows the algorithm to very quickly find and optimize trajectories for multi-robot systems.
Kinodynamic Adaptive Robot Coordination (K-ARC)~\cite{qin2025karc} plans decoupled robot paths and then resolves inter-robot conflicts through solving local subproblems in the composite-robot state space.
To address kinodynamic constraints, the sampled paths are segmented into even lengths, optimized using trajectory optimization, and each segment is incrementally checked for collision to minimize wasted work if a collision occurs. 

All of these methods present effective ways to plan multi-robot motion planning problems with kinodynamic constraints, however, none of them take advantage of guiding information available from the workspace.

\subsection{Workspace Guided Motion Planning}
Existing single and multi-robot planners use workspace guidance to provide sampling guidance, biasing the planner toward specific regions of the configuration space. Several multi-robot planners also use workspace representations to provide coordination guidance, informing the planner when to compose robots into the same configuration space.

\subsubsection{Sampling Guidance}
Prior work has explored leveraging workspace topology to guide single robot motion planners. Several methods use workspace skeleton representations of the environment. Dynamic Region-biased RRT (DR-RRT)~\cite{denny2020drrrt} and its variants~\cite{sandstrom2020drprm,uwacu2022has} create dynamic sampling regions that advance along skeleton edges to guide the planner through narrow passages in the workspace.
In the multi-robot domain, Composite Dynamic Region-biased RRT (CDR-RRT)~\cite{mcbeth2023cdrrrt} extends DR-RRT to coupled multi-robot motion planning and biases sampling in the composite space.
These approaches work well in environments with many narrow passages. When used in environments without narrow passages; however, skeleton guidance still constricts robot motion to follow the skeleton, which can exacerbate risk of collisions in the multi-robot case.

The Synergistic Combination of Layers of Planning (SyCLoP) approach~\cite{plaku2010syclop} proposes a hierarchical single robot planner wherein a region decomposition is used to model the workspace. Using discrete search, a high-level path over the regions for the robot is found over this decomposition. Then, this region path is used to guide construction of a path through configuration space. During each iteration of the algorithm, a region along this high-level path is chosen using a heuristic that takes into account current progress in covering each region and the freespace volume of each region. A sample is generated from the selected region and used to grow an RRT through configuration space. The decomposition region representation used by SyCLoP is not dependent on specific features such as narrow passages existing in the environment which make skeleton-based planners advantageous and still provides effective guidance in open environments where skeletons do not.
Additionally, the hierarchical structure of the algorithm lends itself well to extension to the multi-robot motion planning problem as we can reason about inter-robot coordination using these high-level paths.

\subsubsection{Coordination Guidance}
Workspace Guided-DaSH (WG-DaSH)~\cite{mcbeth2023hypergraph} and its kinodynamic variant build off of CDR-RRT and proposes a hybrid method that uses skeleton guidance both to guide the planner through narrow passages and to inform the planner when coordination between robots is needed.
It constructs abstract paths for each robot over the skeleton using multi-agent pathfinding and then composes the robots into the same configuration space when they are traversing the same region of the skeleton.
Open or heterogeneous environments are not well represented by skeletons because guidance along skeleton edges restricts the planner to a narrow portion of the freespace around the skeleton.
This often forces multiple robots into the same area of the workspace, leading the coordination guidance to compose several robots into the same configuration space that could otherwise be decomposed, posing a computational burden.
In this work, we build off of the idea of using topology to provide insights about coordination between robots during planning and additionally reason over our workspace representation to reduce the amount of coordination needed, expediting planning.

\begin{table}[]
\centering
\begin{tabular}{|l|l|l|l|}
\hline
                    & K-ARC     & K-WG-DaSH     & \textbf{CIPHER}                                                                            \\ \hline
Guidance Type       & kinematic & skeleton      & \textbf{cellular decomp}                                                                   \\ \hline
Coordination Guidance  &           & skeleton MAPF & \textbf{cellular decomp MAPF}                                                              \\ \hline
Resolution Guidance &           &               & \textbf{\begin{tabular}[c]{@{}l@{}}subproblem hierarchical\\ cellular decomp\end{tabular}} \\ \hline
\end{tabular}
\caption{Use of Guidance in Multi-robot Kinodynamic Planning}
\label{tab:guidance-usage}
\vspace{-20pt}
\end{table}

\section{Proposed Method}
In this section, we present Coordinated Incremental Planning with Hierarchical Expansion and Refinement (CIPHER), which leverages the use of topological guidance, similar to SyCLoP~\cite{plaku2010syclop}, with resolution guidance to enable coordination provided by a MAPF layer and region-guided planners.

\subsection{Problem Formulation}
We consider the problem of planning conflict-free trajectories for $n$ robots operating in a shared two-dimensional workspace $W \subseteq \mathbb{R}^2$ containing $m$ static obstacles $\mathcal{O} = \{ O_1, ...,O_m\}$.
Each robot $r_i$ is represented by a state $x_i \in \mathcal{X}_i$, where $\mathcal{X}_i$ denotes the state space of robot $r_i$.
Each robot begins in a start state, $s_i \in \mathcal{X}_i~\forall_i \in [1,n]$, and must traverse to a goal state, $g_i \in \mathcal{X}_i~\forall_i \in [1,n]$.
We assume that each robot has dynamics $\dot{x}_i = f_i(x_i, u_i)$ where $u_i \in \mathcal{U}_i$ is a control input for the robot in the robots admissible control set. 

The multi-robot motion planning problem seeks to find a set of dynamically feasible trajectories $\{\tau_1,...,\tau_n\}$, where $\tau_i$ is a trajectory for robot $r_i$, such that:
\begin{enumerate}
    \item Each trajectory connects its start to goal: $\tau_i(0) = s_i$ and $\tau_i(T_i) = g_i$ where $T_i$ is the last timestep in trajectory of robot $i$.
    \item At any time $t$ during execution, the state $(x_1(t), ..., x_n(t))$ is inter-robot collision-free. We extend trajectories beyond their terminal time by assuming that robots remain stationary at their goals.
\end{enumerate}

\subsection{Naive Resolution Guidance Planner}
A natural baseline for incorporating workspace guidance into multi-robot motion planning is to combine naive resolution guidance, where only a single resolution is considered, with decoupled continuous space planning. The workspace is partitioned into grid cells, as in SyCLoP~\cite{plaku2010syclop}, and a multi-agent path finding (MAPF) algorithm such as CBS computes a conflict-free sequence of region transitions for each robot.
Each robot then independently runs a guided RRT using its assigned region path as a sampling heuristic and setting higher priority planned robot paths as dynamic obstacles.
We name this baseline Prioritized Planning Region-Guided RRT (PP-RG-RRT).

The decoupled behavior allows for cheap coordination using MAPF.
However, the MAPF performed only guarantees that the region sequences are collision free during the high level search. 
During the continuous planning, the paths of higher priority robots may block the paths of lower priority robots within cells.
This baseline does not contain conflict detection or resolution behavior which can lead to failures as the robot density or decomposition resolution increases.
Our proposed method CIPHER retains this baselines two-phase structure but augments it with explicit conflict detection and a hierarchical resolution phase that targets exactly this gap.

\subsection{CIPEHR Overview}
CIPHER is a hierarchical planning framework that leverages the use of resolution guidance during multi-robot motion planning.
The approach follows a plan-check-resolve paradigm where independent robot plans are initially created using region-level topological guidance. Then the trajectories are checked for inter-robot collisions and detected conflicts are resolved through increasingly aggressive conflict resolution strategies starting at the region-level.

The framework utilizes insights from our prior work~\cite{mcbeth2023hypergraph}, demonstrating the benefit of topological guidance in multi-robot motion planning, to extend the single-robot SyCLoP planner~\cite{plaku2010syclop}, which combines discrete search over workspace decompositions with continuous sampling-based planning, to the multi-robot setting.

The framework consists of four phases as outlined in Algorithm~\ref{alg:cipher}:
\begin{enumerate}
    \item \textbf{Phase 1 (Decomposition \& MAPF):} Partition the workspace into grid cell regions and compute conflict-free paths of regions via multi-agent path finding (MAPF). The MAPF algorithm will find coordinated high-level guiding paths at the workspace level.
    \item \textbf{Phase 2 (Guided Planning):} Each robot independently computes a dynamically feasible trajectory guided by its computed region path from Phase 1. The region path serves as a heuristic guidance to bias the sampling-based planner towards the regions along the planned region path.
    \item \textbf{Phase 3 (Conflict Detection):} Trajectories are discretized into fixed-duration segments for conflict checking. The discretizations will help in evaluating the inter-robot conflict checks during planning. Check all robot pairs for geometric conflict at each timestep along the trajectory segments.
    \item \textbf{Phase 4 (Conflict Resolution):} Resolve any detected conflicts through a set of escalating hierarchical strategies: region cell decomposition refinement, region expansion, and falling back to composite planning if all strategy attempts fail.
    \item \textbf{Phase 5 (Fallback Planning):} If the conflicts cannot be resolved, the planner falls back to a composite planner to solve the whole problem. 
\end{enumerate}

\begin{algorithm}[t]
\small
    \caption{CIPHER Main Loop}
    \label{alg:cipher}
    \begin{algorithmic}[1]
        \INPUT Robots $\{r_i\}_{i=1}^{n}$ with $(s_i, g_i)$, obstacles $\mathcal{O}$, Workspace $\mathcal{W}$ and parameters $l$
        \OUTPUT Conflict-free trajectories $\tau = (\tau_1,...,\tau_n)$ or $FAILURE$
        \State $D \gets GridDecompose(\mathcal{W}, l)$
        \State $(P_1,...,P_n) \gets MAPF(D, \{s_i\}, \{g_i\})$
        \For{$i = 1$ to $n$}
            \State $\tau_i \gets GuidedPlan(r_i, s_i, g_i, D, P_i)$
        \EndFor
        \State $C \gets DetectConflicts(\{\tau_i\})$
        \While{$C \neq \emptyset$}
            \State $c \gets Pop(C)$
            \If{not $ResolveConflicts(c, D)$}
                \State break;
            \EndIf
            \State $C \gets RecheckConflicts(\{\tau_i\})$
        \EndWhile
        \If{$C \neq \emptyset$}
            \State $\{\tau_i\}_{i=1}^{n} \gets FallbackPlanner(\{r_i\}_{i=1}^{n}, \{s_i\}_{i=1}^{n}, \{g_i\}_{i=1}^{n})$
        \EndIf
        \State return $(\tau_1, ..., \tau_n)$
    \end{algorithmic}
\end{algorithm}

\begin{algorithm}[t]
\small
    \caption{$ResolveConflicts$}
    \label{alg:resolve}
    \begin{algorithmic}[1]
        \INPUT Conflict $c = (r_a, r_b, t)$, Decomposition $D$
        \OUTPUT $true$ if $c$ is resolved, $false$ if $c$ is not resolved
        \State $R \gets LocateRegion(c)$
        \State $L_{\min} \gets ExpansionLayer(c)$
        \For{$L = L_{\min}$ to $MaxExpansion(D)$}
            \State $R_{exp} \gets ExpandRegion(R, L)$
            \For{$k = 1$ to $MaxRefinement$}
                \If{$RefineAndReplan(c, R_{exp}, k, D)$}
                    \State \Return $true$
                \EndIf
            \EndFor
            \If{$R_{exp}$ covers $D$} \textbf{break} \EndIf
        \EndFor
        \State \Return $false$
    \end{algorithmic}
\end{algorithm}

\subsection{Workspace Decomposition}
The workspace $\mathcal{W}$ is partitioned into a uniform grid of rectangular cells following the approach presented in SyCLoP~\cite{plaku2010syclop}.
We compute the obstacle occupancy percentage of each decomposition cell and mark cells above a specified threshold as occupied.
This forms a decomposition of $N = l^2$ non-overlapping cells, where $l$ is the grid resolution, which provides a discrete abstraction of the continuous workspace that we use for high-level reasoning about robot coordination.

A region graph, represented as an adjacency graph, $G_D = (V_D, E_D)$ is defined, where a vertex $v \in V_D$ represents the decomposition cells, $c_i \in V_D~\forall_i \in \{0,...,N-1\}$, and an edge $(v,w) \in E_D$ contains the 4-connected cardinal neighbors of each respective cell.
A projection function $\phi: \mathcal{X}_i \to V_D$ maps the robot states to region cells by extracting position information and identifying the corresponding region cell that contains it.

The grid-based decomposition offers some advantages over the skeleton-based representation used in prior work~\cite{mcbeth2023hypergraph}.
The grid decomposition naturally covers the entire workspace, which provides guidance even in open environments where the skeleton decompositions would restrict planning to narrow corridors.
Additionally, our insight is that the grid cells can be refined to increase resolution in areas where conflicts occur, allowing us to explore the use of grid refinement as a conflict resolution technique. 

\subsection{High-level Planning via Multi-Agent Path Finding}
The high-level planning layer computes a coordinated path of regions for all robots by solving a Multi-Agent Path Finding (MAPF) problem on the region graph $G_D$.
To reduce congestion along the region paths and the need to decompose cells, regions are given a capacity limit.
Each robot $r_i$ has a start region $\phi(s_i)$ and goal region $\phi(g_i)$.
The output of the high-level plan is a set of region paths $P_i = (c_0^i, c_1^i,...,c_{T_i}^i)$ that guide each robot from its start to its goal region.

Note that this high-level planning layer operates on the region graph and is thus independent of any robot-specific information, such as dynamics.
This enables the use of efficient MAPF algorithms that scale well with the number of robots such as Conflict-Based Search (CBS)~\cite{SHARON201540}.
CBS provides optimal solutions by resolving conflicts using a two-level search which has been shown to be very effective at solving MAPF problems with lots of coordination.

The output of the MAPF layer provides workspace-level coordination where robots are assigned paths through the region graph that avoid occupying the same cell as other robots at the same timestep.
While this coordination does not guarantee conflict-free trajectories, it provides guidance that may be useful in reducing the likelihood of conflicts in the composite-space resolution.

\subsection{Region-Guided Low-Level Planning}
Utilizing the region paths $P_1, ..., P_n$ from the high-level planning layer, the low-level planning layer computes dynamically feasible trajectories $\{\tau_i\}_{i=1}^{n}$ for each robot $r_i$ independently.
Each trajectory must connect $s_i$ to $g_i$ while approximately following the sequence of regions in $P_i$ and avoiding obstacles.

\subsubsection{Guided Planning}
The guided planning layer leverages a planning interface where, given a robot $r_i$, a decomposition $D$, a region path $P_i$, the robots start $s_i$ and goal $g_i$ and returns a dynamically feasible trajectory $\tau_i$.
The region path $P_i$ serves as a heuristic guide, biasing the underlying planners search towards regions in $P_i$, but not strictly constrained to them, preserving the probabilistic completeness of the underlying sampling-based method.
Unlike K-ARC and K-WG-Dash, which utilize waypoints or path segments to guide exploration, respectively, CIPHER utilizes the region-based guidance similar to SyCLoP~\cite{plaku2010syclop} to grow a tree that biases samples towards the region path $P_i$.
Similar to SyCLoP, we keep a probability of sampling from the environment rather than biasing towards the region path.
Unlike SyCLoP, we don't keep track of failed extensions to specific regions as we rely on our conflict resolution strategy to find valid paths.
We then segment the path and check for conflicts afterwards.

We provide two implementations: Guided RRT Connect for geometric systems and Region-Guided DB-RRT for kinodynamic systems.

\textbf{Geometric Case: Guided RRT Connect}
For robots with relatively simple dynamics, such as point robots, we utilize a variant of RRT which biases samples to follow the high-level MAPF guidance.

\textbf{Kinodynamic Case: Region-Guided DB-RRT}
When robots have differential constraints, we leverage the work of iterative Discontinuity Bounded RRT (iDb-RRT)~\cite{o-idbrrt-24} while incorporating region guidance.
The iDb-RRT algorithm samples sequences of motion primitives with bounded discontinuity and then applies trajectory optimization to repair the discontinuities and produce dynamically feasible trajectories.

We incorporate region guidance by biasing the sampling distribution towards the cells in the region path $P_i$. Specifically, when selecting a target for tree extension, we weight cells according to their position along the region path.
This biasing directs the kinodynamic planner towards the routes computed by the MAPF solver still allowing the motions to satisfy dynamic constraints.

\begin{figure}
    \vspace{-10pt}
    \centering
    \hfill
    \subfigure[Decomposition Refinement]{
        \includegraphics[width=0.2\textwidth]{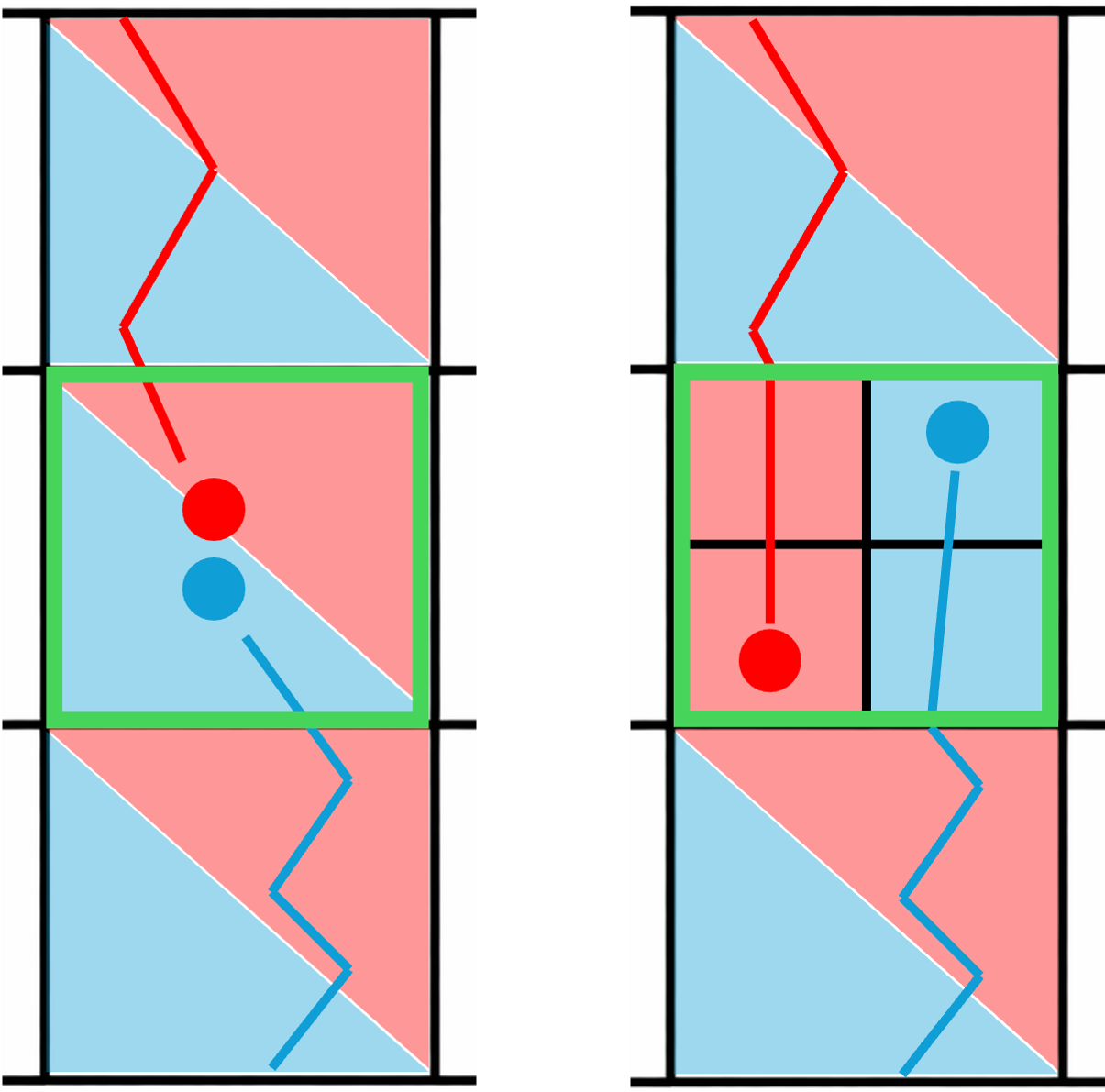}
    }
    \hfill
    \subfigure[Spatial Expansion]{
        \includegraphics[width=0.4\textwidth]{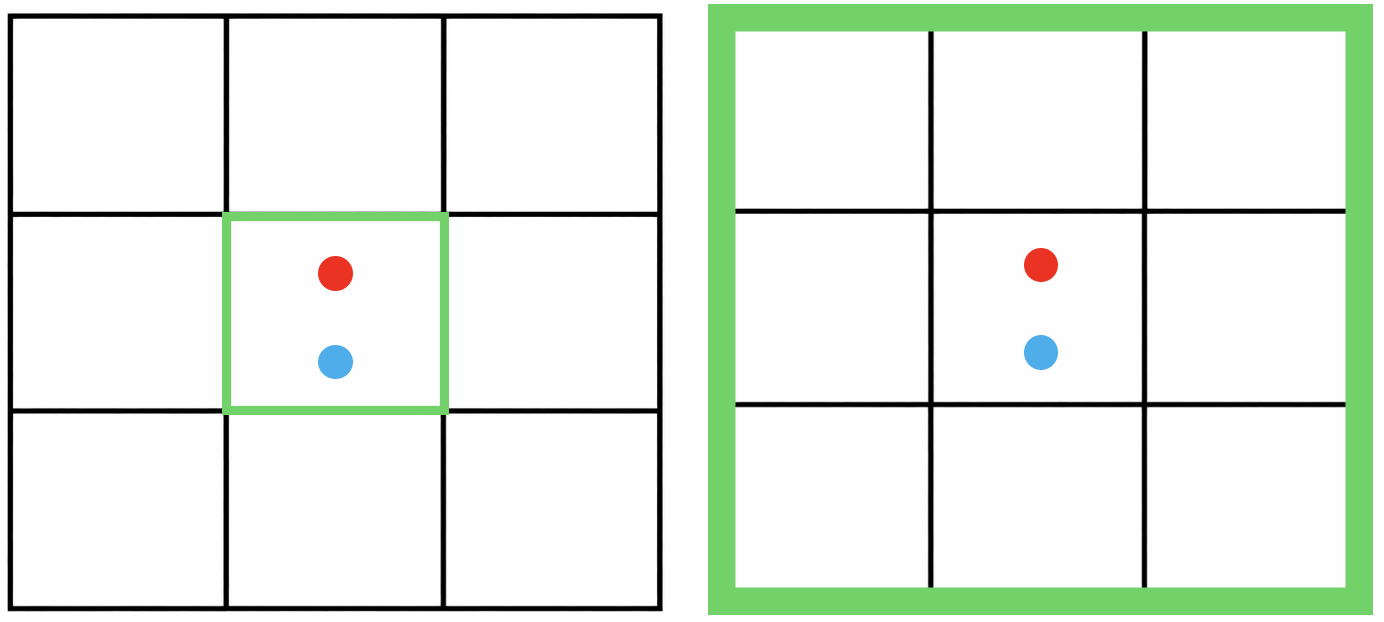}
    }
    \hfill
    \hfill
    \caption{Conflict Resolution: (a) depicts the decomposition refinement strategy (resolution guidance) which refines the conflict cell into a finer resolution. (b) depicts the spatial expansion strategy where the region being considered during MAPF is expanded to a larger area.}
    \label{fig:conflict-resolution}
    \vspace{-25pt}
\end{figure}

\subsection{Hierarchical Conflict Resolution}
When conflicts are detected, our method resolves them through an escalating hierarchy of adaptive resolution guidance strategies, modifying only the colliding robots trajectories while preserving others as described in Algorithm~\ref{alg:resolve}.
\subsubsection{Strategy 1: Decomposition Refinement} 
The first resolution guidance strategy refines the workspace decomposition locally around the conflict, as seen in Figure~\ref{fig:conflict-resolution} (a). We refine the decomposition by decomposing the cell in which the conflict is detected in.
Using the projection function, $\phi$, we can find the conflict state $x_{i}^{t_c}$ and find the associated region cell in which the conflict occurs using $R_c = \phi(x_{i}^{t_c})$.
The insight from this is that conflicts may resolve when finer decompositions are utilized in the conflict region, which provides the MAPF solver additional route options that may lead robots to traverse different sub-regions, potentially avoiding conflict without requiring composite space planning.

After refinement, the MAPF problem is re-solved in the refined conflict cell for the robots involved in the conflict and new guided trajectories are computed.
If conflicts persist until the cell cannot be decomposed any further, cells must always be larger than robot size, the algorithm proceeds to the next strategy.

\subsubsection{Strategy 2: Spatial Expansion}
If the decomposition refinement fails to resolve the conflict, we expand the refinement region to include all the cells adjacent to the conflict cells, including the diagonal neighbors, as seen in Figure~\ref{fig:conflict-resolution} (b).
The expanded region is then subdivided, providing additional flexibility in routing the robots in a larger neighborhood around the conflict.
The spatial expansion strategy addresses the cases where the conflict cannot be resolved within a single cell.
The refinement of the larger area allows the planner to discover alternative routes through different areas of the expanded region.

Similar to the decomposition refinement, after expanding and refining, the MAPF problem is re-solved in the expanded conflict cells for the conflicting robot, followed by the guided trajectory computation.
The expansion can be iteratively increased, starting from a minimum set of regions $L_{min}$, if the conflicts persist, eventually encompassing the entire workspace.

\subsection{Fallback Planning}
If the entire hierarchical resolution finishes without resolving a conflict, CIPHER abandons region guidance and reverts to standard multi-robot planners.
We utilize 2 fallback planners in this work.
First, we leverage a prioritized planning decoupled RRT planner, where robots plan independently treating preceding robots as dynamic obstacles, as it tends to be fast and reliable for the types of problems we are considering.
If decoupled planning fails to find a solution within a set time limit, we utilize a composite RRT which plains in the joint configuration space of all robots.
The composite planner is probabilistically complete but scales exponentially with the number of robots, so it is reserved for problems where every cheaper strategy has been exhausted.



\section{Analysis}

\textit{Theorem:} CIPHER is probabilistically complete. \\
\textit{Proof:} 
Whenever its conflict resolution strategy fails, CIPHER falls back to a composite RRT planner~\cite{db-spsanomp-13} which samples directly in the joint configuration space of all the robots and is probabilistically complete for the multi-robot motion planning problem.
Since every execution of CIPHER that does not find a solution from the hierarchical procedure falls back to the composite planner and the composite planner is probabilistically complete, the probability that CIPHER returns a valid set of robot trajectories on a feasible problem instance approaches one as time increases. Thus, CIPHER is probabilistically complete.



\section{Experimental Validation}
We run scaling experiments to compare the performance of CIPHER and PP-RG-RRT against other state-of-the-art multi-robot motion planning methods. We compare the geometric implementations of CIPHER and PP-RG-RRT to Coupled RRT~\cite{sl-uppccdpmrs-2002} which runs an RRT in the full composite space of the robot group, Decoupled RRT~\cite{bennewitz2001prioritized} which plans for each robot separately using prioritized planning, Subdimensional-expansion RRT (sRRT)~\cite{wagner2012srrt}, Adaptive Robot Coordination (ARC)~\cite{solis2024arc}, Workspace Guided-DaSH (WG-DaSH)~\cite{mcbeth2023hypergraph}, and Discrete RRT (MRdRRT)~\cite{solovey2016drrt}. We compare the kinodynamic implementations of CIPHER and PP-RG-RRT to Coupled Kinodynamic RRT~\cite{sl-uppccdpmrs-2002}, Decoupled Kinodynamic RRT~\cite{bennewitz2001prioritized}, Kinodynamic ARC (K-ARC)~\cite{qin2025karc} and Kinodynamic WG-DaSH~\cite{mcbeth2023hypergraph}.

By comparing to Coupled and Decoupled RRT, we demonstrate the impact of using topological guidance to decide when coordination is needed rather than using a fixed level of coordination during planning. Comparisons to sRRT, ARC, and MRdRRT show the benefits of using guidance to proactively avoid inter-robot conflicts rather than reactively responding to conflicts. The comparison to WG-DaSH demonstrates that CIPHER supports a broader range of environment topologies and shows the impact of refining the workspace decomposition.

Our implementation of CIPHER is publicly available at \url{https://github.com/parasollab/cipher}. We implemented our approach as well as the comparison methods in the Open Motion Planning Library~\cite{sucan2012the-open-motion-planning-library} with the exception of WG-DaSH, which is publicly available in the Parasol Planning Library~\cite{mcbeth2023hypergraph}. Experiments were run on a computer with an Intel Core Ultra 7 265K CPU.

Our experimental environments are described below. Each method is given 600 seconds to solve each scenario. We run scaling problems with multiples of 2 robots up to 32 robots or until all ten seeds for each method fail. Starts and goals are randomly generated. For kinodynamic planning methods, we model the robot using first-order unicycle kinematics with state 
$\mathbf{x} = [x, y, \theta]^\top$ and control inputs 
$\mathbf{u} = [v, \omega]^\top$, given by 
$\dot{x} = v\cos\theta,\ \dot{y} = v\sin\theta,\ \dot{\theta} = \omega.$

\begin{figure}[t]
    \centering
    \hfill
    \subfigure[Empty]{
        \includegraphics[width=0.17\textwidth]{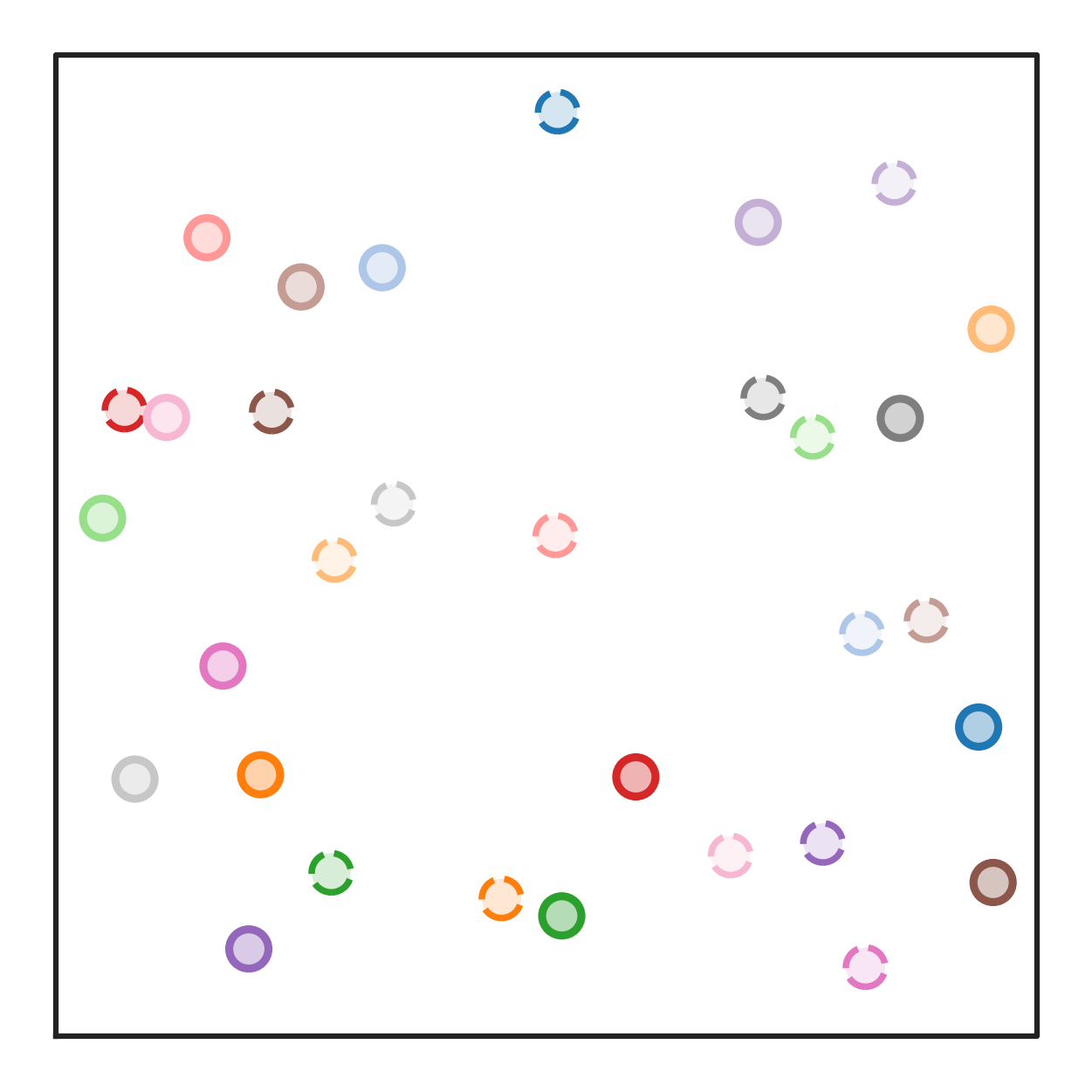}
        \label{fig:empty-env}
    }
    \hfill
    \subfigure[Rooms]{
        \includegraphics[width=0.17\textwidth]{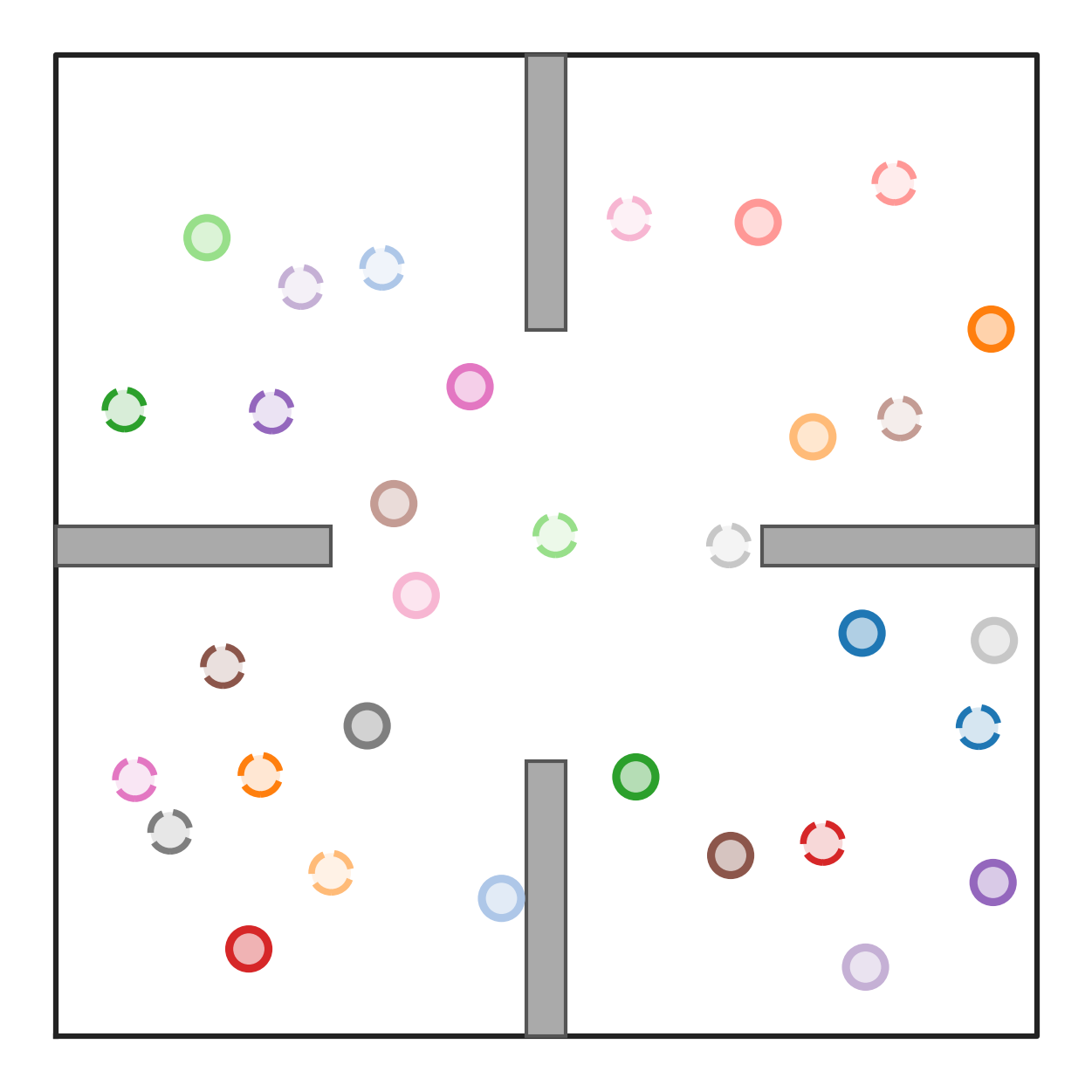}
        \label{fig:rooms-env}
    }
    \hfill
    \subfigure[Low]{
        \includegraphics[width=0.17\textwidth]{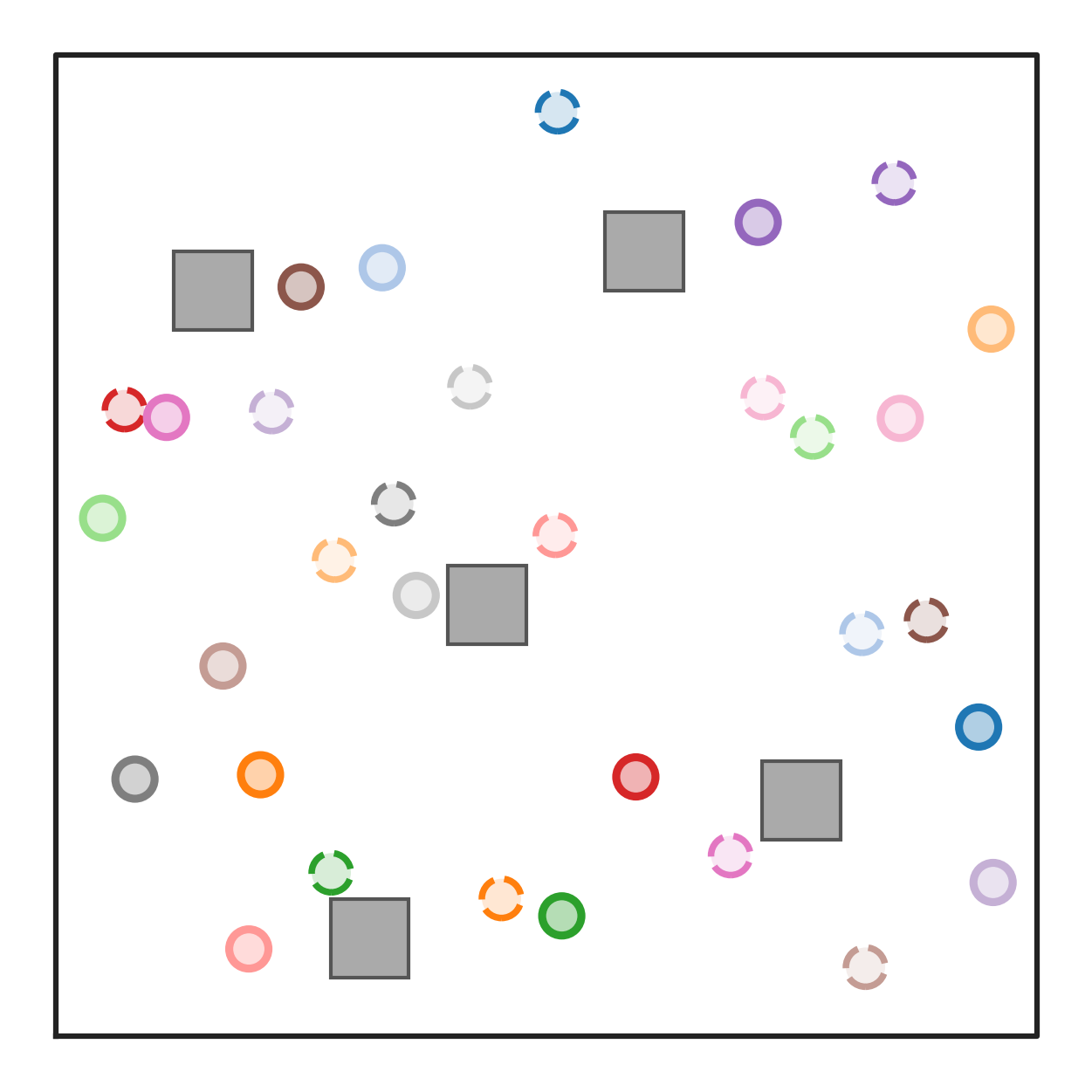}
        \label{fig:low-env}
    }
    \hfill
    \subfigure[Medium]{
        \includegraphics[width=0.17\textwidth]{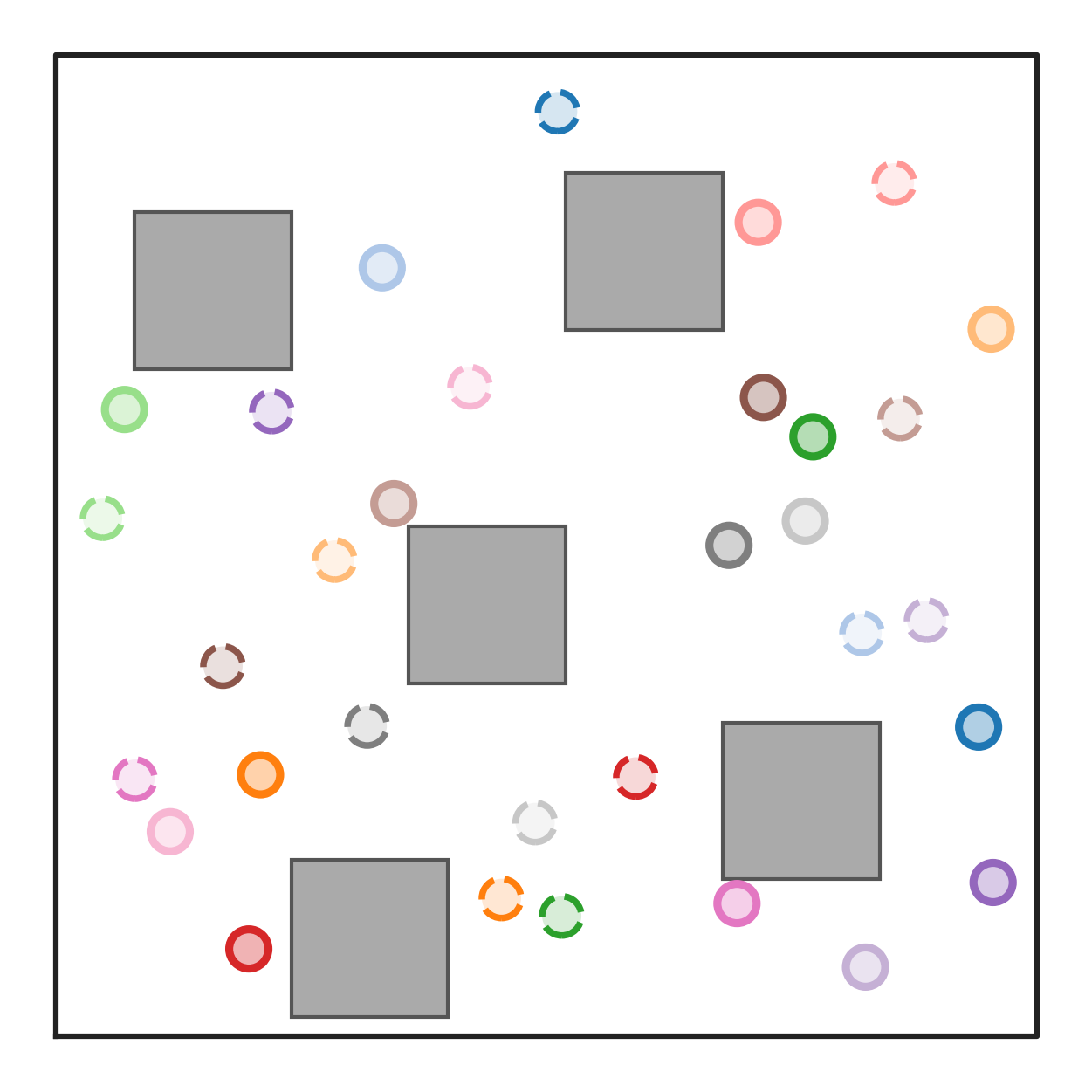}
        \label{fig:medium-env}
    }
    \hfill
    \subfigure[High]{
        \includegraphics[width=0.17\textwidth]{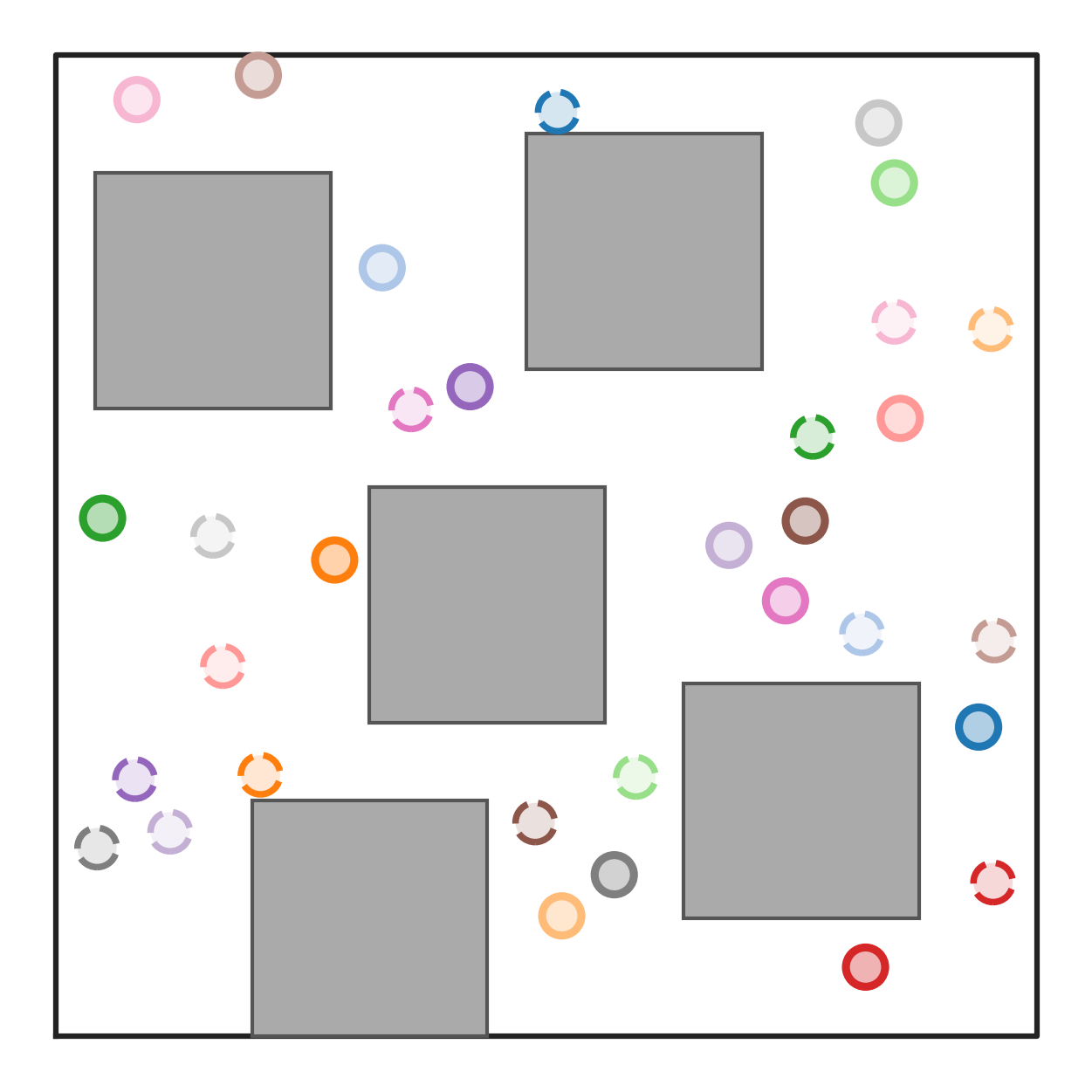}
        \label{fig:high-env}
    }
    \hfill
    \hfill
    \caption{Our experimental environments with starts and goals for a 16-robot problem. Starts and goals are shown as outlines of the robots with goals as dashed lines.}
    \label{fig:envs}
    \vspace{-15pt}
\end{figure}

\subsection{Environments}

\subsubsection{Empty}
The empty environment (Fig.~\ref{fig:empty-env}) is intended to highlight the ability of workspace region-based guidance to support a broad range of topological features. While WG-DaSH is intended for environments featuring narrow passages, our method can utilize guidance even when no obstacles are present.

\subsubsection{Rooms}
This environment (Fig.~\ref{fig:rooms-env}) features walls that separate open spaces, forcing robots through a congested space to transition between rooms. We show the ability of CIPHER to accommodate a diverse set of features in the environment, while other methods may struggle with other aspects of the environment.

\subsubsection{Random Obstacles}
We use three environments with randomly distributed obstacles of different sizes. These environments (Figs.~\ref{fig:low-env},~\ref{fig:medium-env}, and~\ref{fig:high-env}) scale from low obstacle volume to high obstacle volume in the workspace, increasing the robot congestion that occurs as the obstacles increase in size, which is especially difficult for kinodynamic planners.
We show our approach's ability to provide coordination by reasoning over the workspace representation.

\subsection{Results and Discussion}
\begin{figure}
    \subfigure[Empty Planning Time]{
        \includegraphics[width=0.45\linewidth]{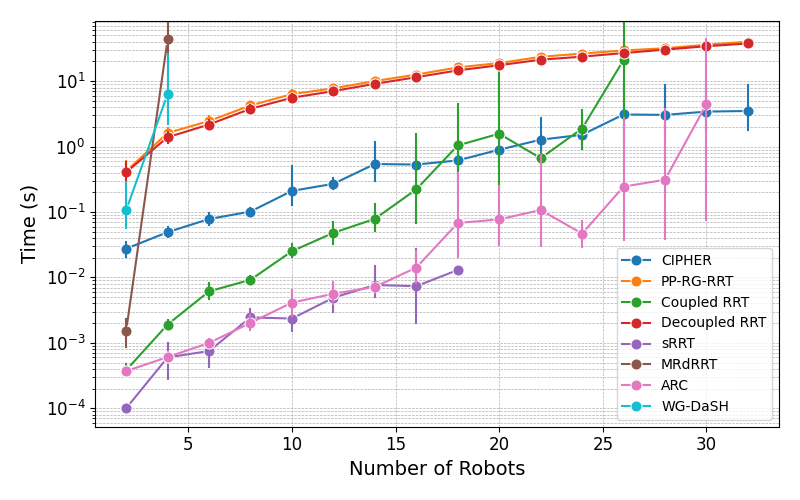}
    }
    \subfigure[Empty Success Rate]{
        \includegraphics[width=0.45\linewidth]{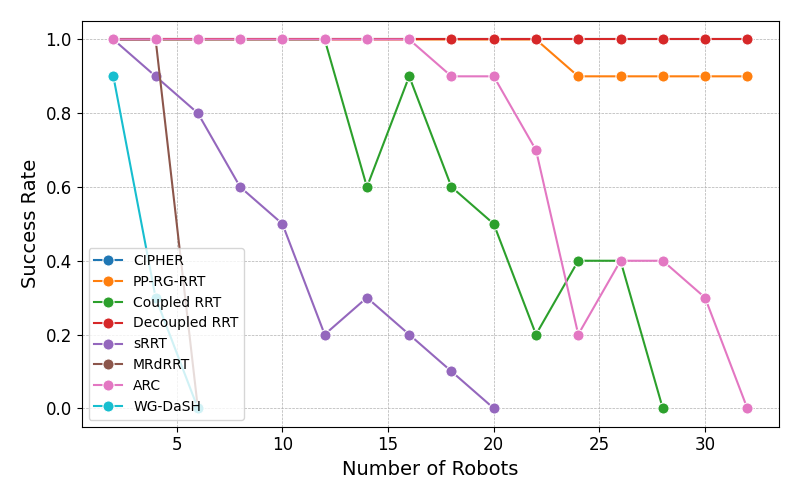}
    }
    \subfigure[Rooms Planning Time]{
        \includegraphics[width=0.45\linewidth]{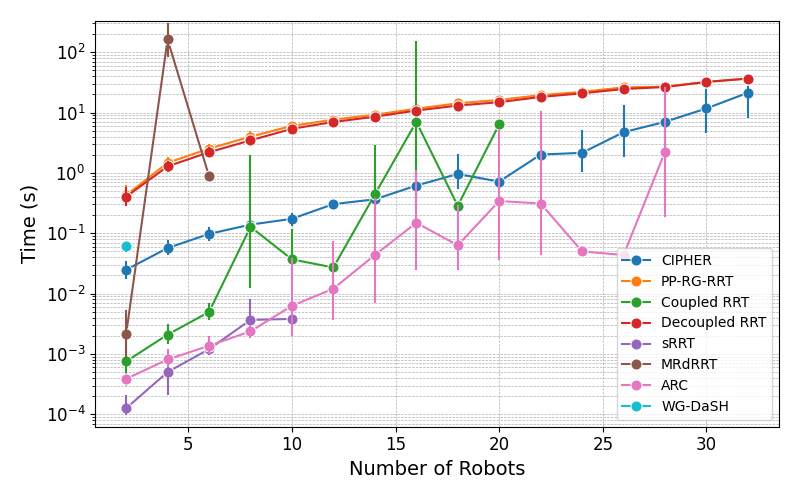}
    }
    \subfigure[Rooms Success Rate]{
        \includegraphics[width=0.45\linewidth]{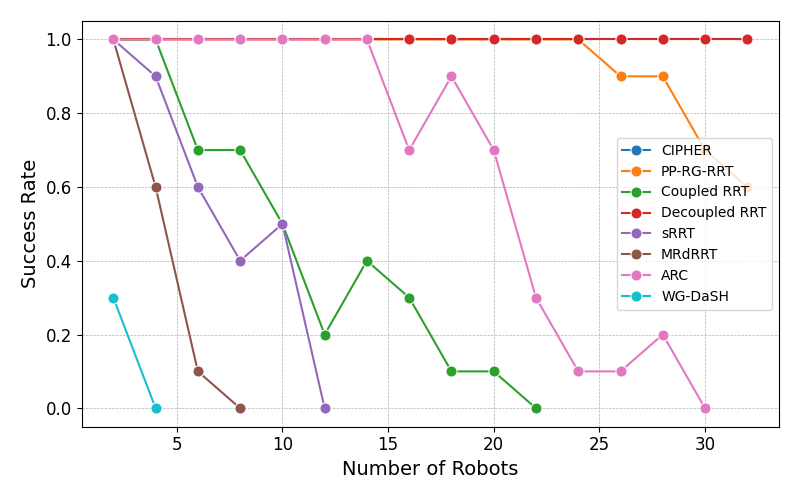}
    }
    \subfigure[Low Clutter Planning Time]{
        \includegraphics[width=0.45\linewidth]{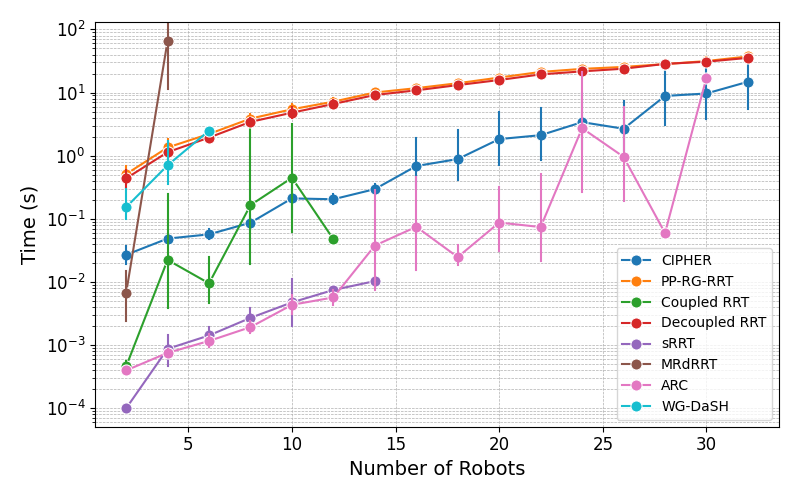}
    }
    \subfigure[Low Clutter Success Rate]{
        \includegraphics[width=0.45\linewidth]{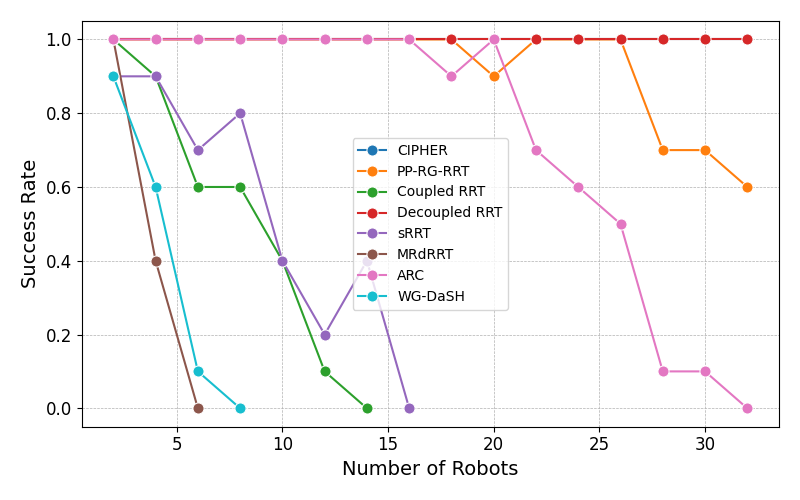}
    }
    \subfigure[Medium Clutter Planning Time]{
        \includegraphics[width=0.45\linewidth]{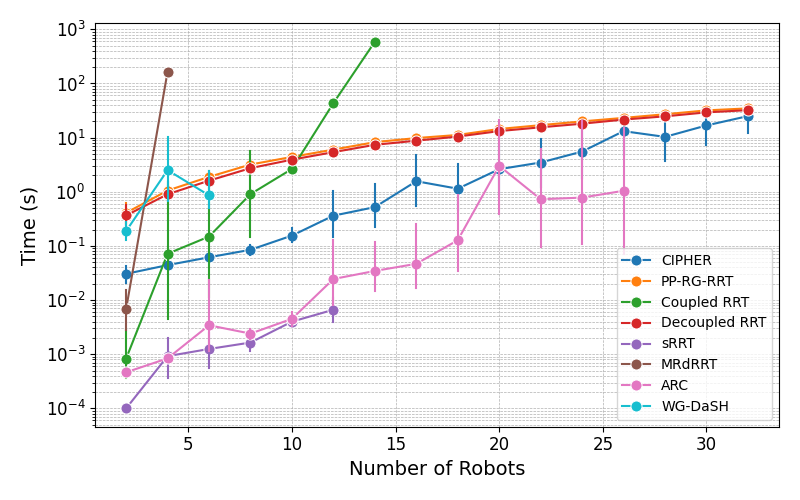}
    }
    \subfigure[Medium Clutter Success Rate]{
        \includegraphics[width=0.45\linewidth]{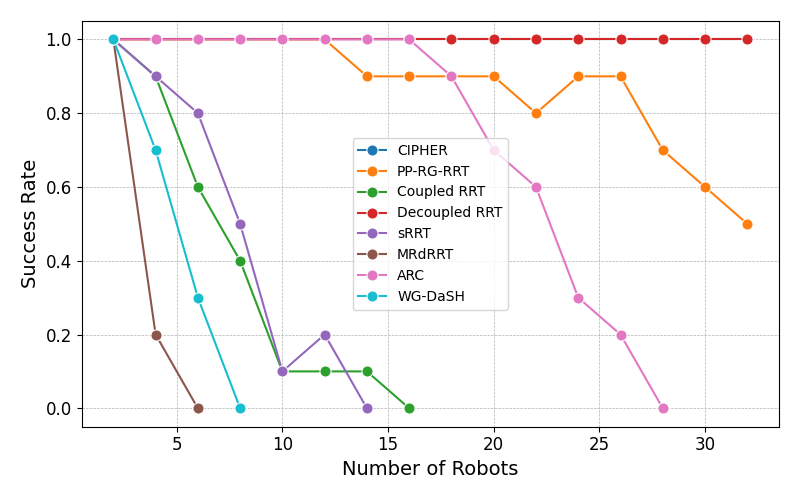}
    }
    \subfigure[High Clutter Planning Time]{
        \includegraphics[width=0.45\linewidth]{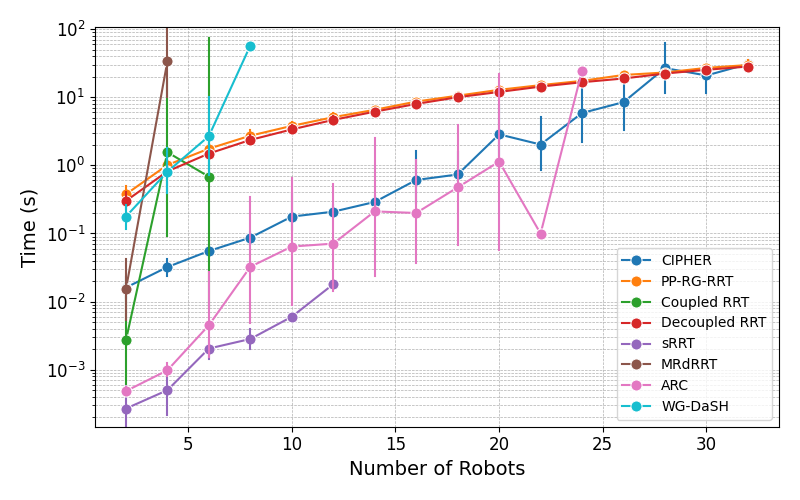}
    }
    \hfill
    \subfigure[High Clutter Success Rate]{
        \includegraphics[width=0.45\linewidth]{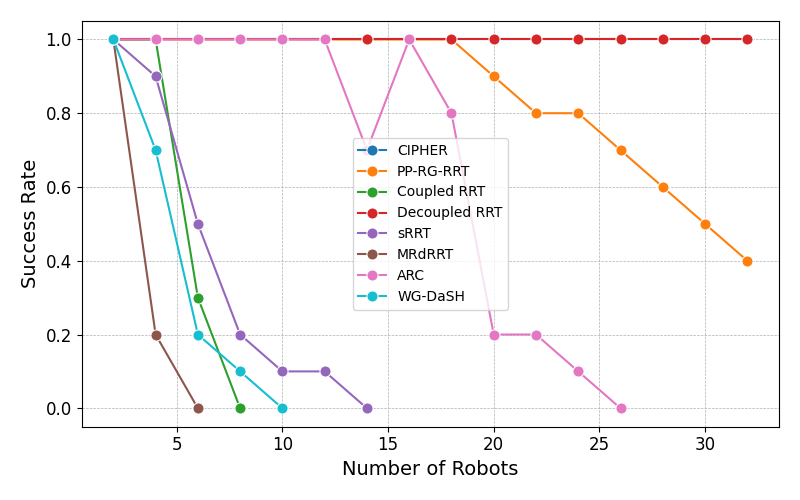}
    }
    
    \caption{Experimental results for the geometric planners showing average planning time and success rate over 10 seeds. Standard deviation in the planning times is shown where more than one seed succeeded. Results omitted where all seeds failed.}
    \label{fig:geometric-results}
\end{figure}

\begin{figure}
    \centering
    \subfigure[Empty Planning Time]{
        \includegraphics[width=0.45\linewidth]{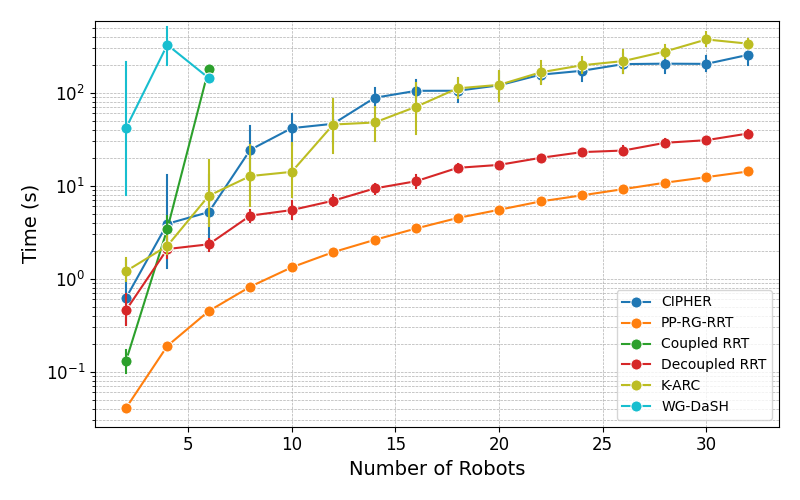}
    }
    \subfigure[Empty Success Rate]{
        \includegraphics[width=0.45\linewidth]{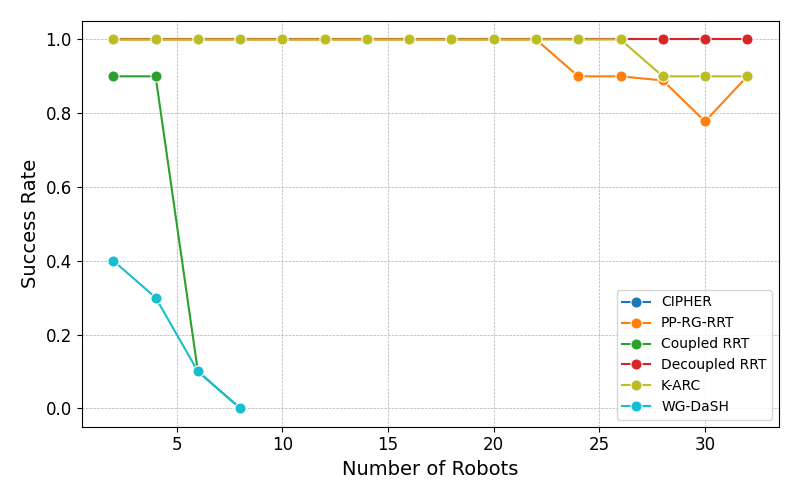}
    }
    \subfigure[Rooms Planning Time]{
        \includegraphics[width=0.45\linewidth]{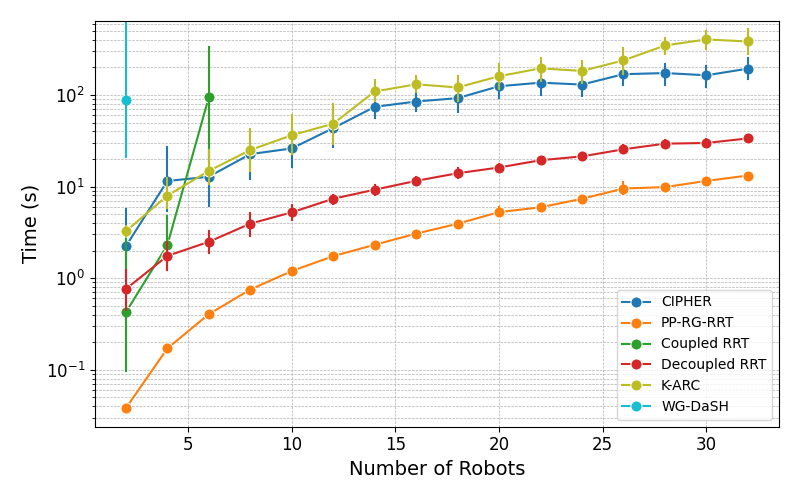}
    }
    \subfigure[Rooms Success Rate]{
        \includegraphics[width=0.45\linewidth]{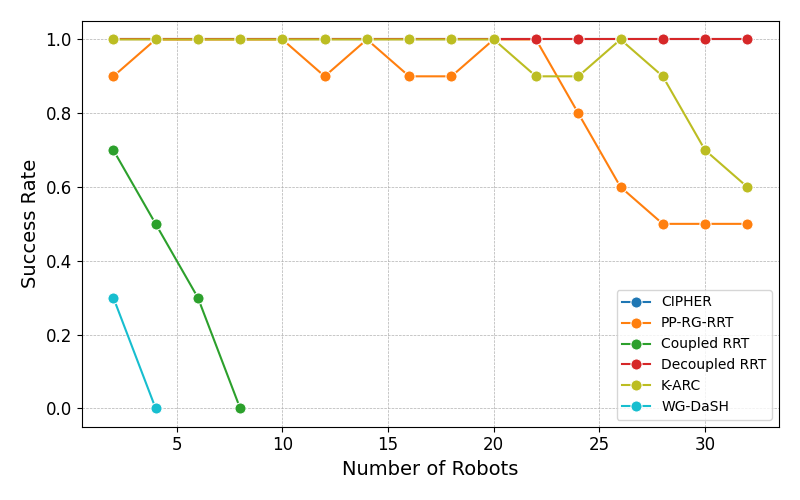}
    }
    \subfigure[Low Clutter Planning Time]{
        \includegraphics[width=0.45\linewidth]{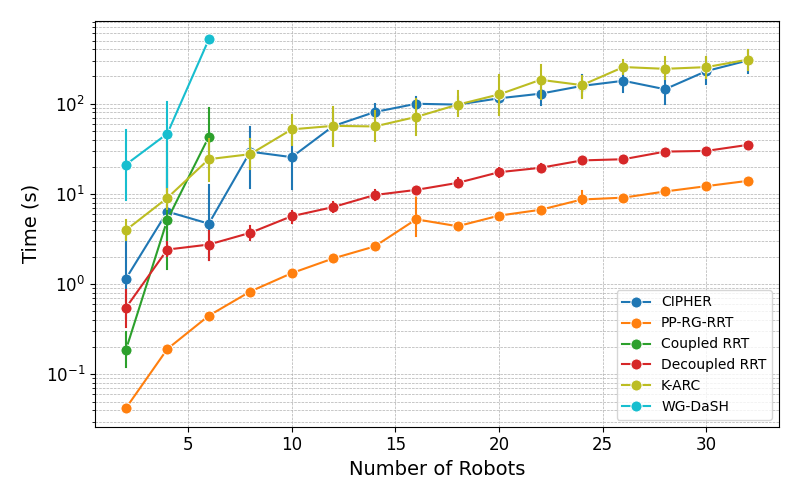}
    }
    \subfigure[Low Clutter Success Rate]{
        \includegraphics[width=0.45\linewidth]{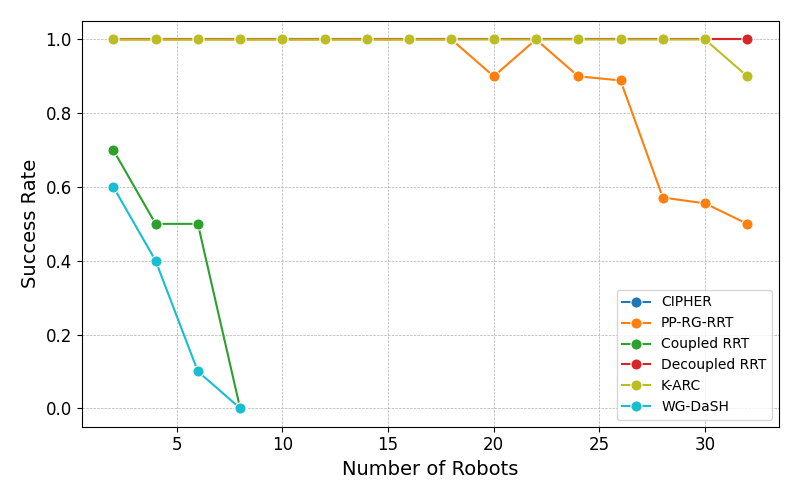}
    }
    \subfigure[Medium Clutter Planning Time]{
        \includegraphics[width=0.45\linewidth]{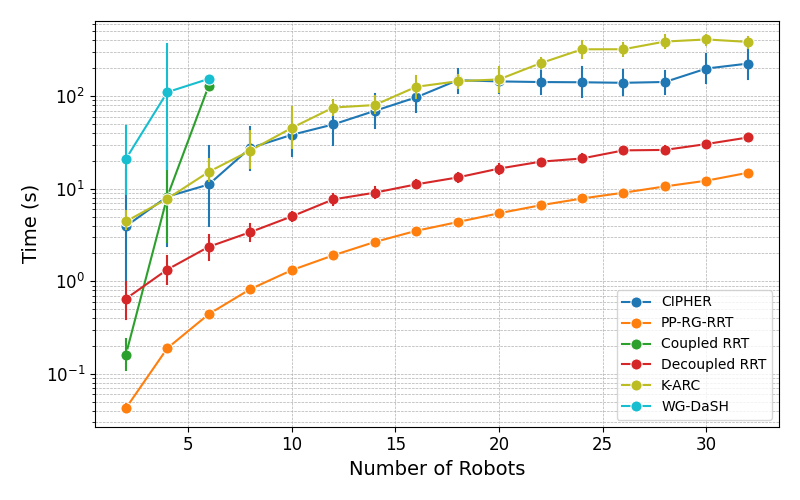}
    }
    \subfigure[Medium Clutter Success Rate]{
        \includegraphics[width=0.45\linewidth]{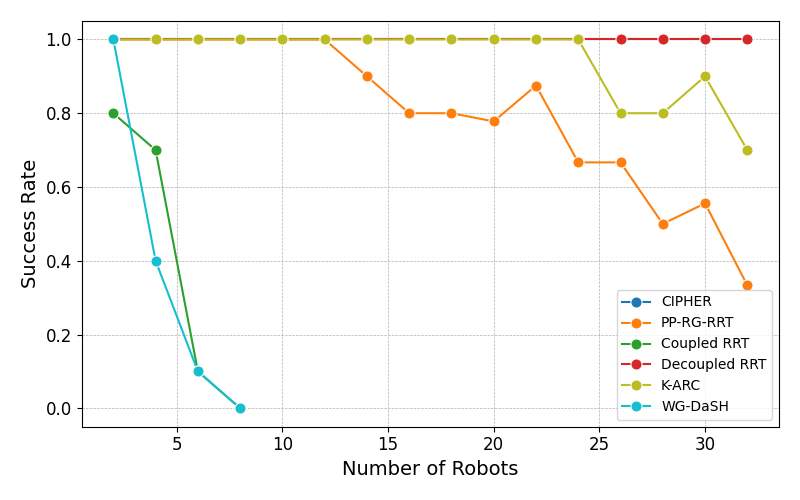}
    }
    \subfigure[High Clutter Planning Time]{
        \includegraphics[width=0.45\linewidth]{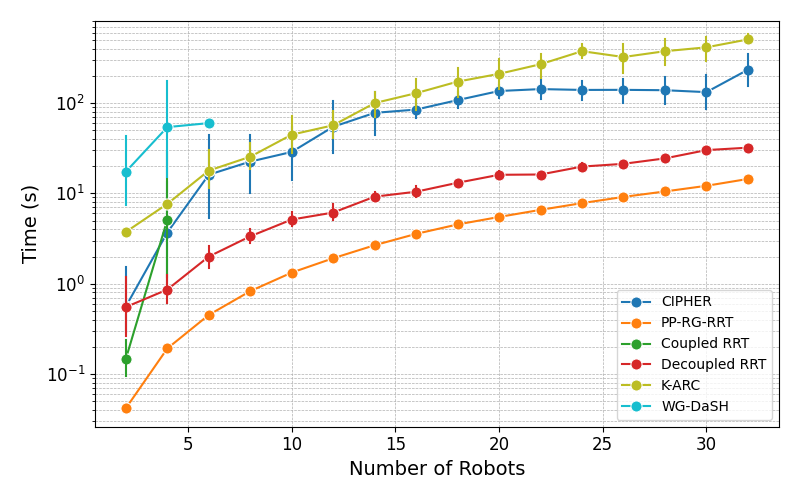}
    }
    \subfigure[High Clutter Success Rate]{
        \includegraphics[width=0.45\linewidth]{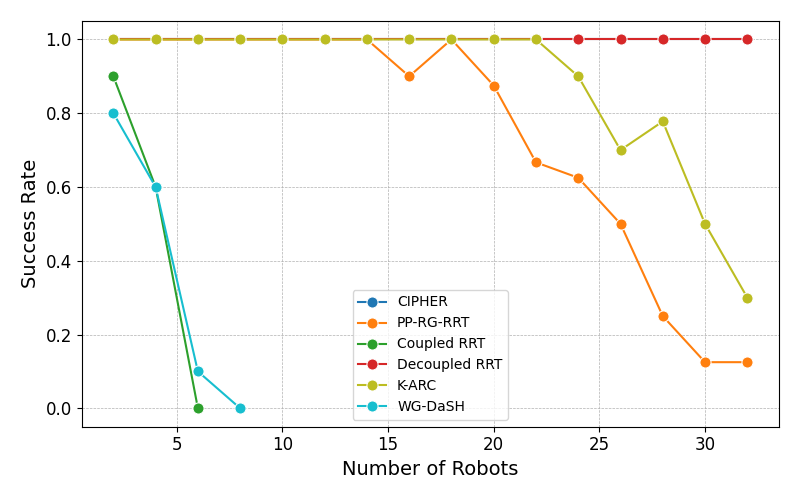}
    }
    \caption{Experimental results for the kinodynamic planners showing average planning time and success rate over 10 seeds. Standard deviation in the planning times is shown where more than one seed succeeded. Results omitted where all seeds failed.}
    \label{fig:kinodynamic-results}
\end{figure}

\subsubsection{Geometric Planners}
Results for the geometric planners are shown in Fig.~\ref{fig:geometric-results}. In all environments, CIPHER is able to solve all problems with a 100\% success rate. The only other method that is able to achieve this is Decoupled RRT, which performs well in open environments; however, CIPHER achieves up to an order of magnitude lower planning times. Coupled RRT and MRdRRT suffer as a result of searching large spaces. WG-DaSH uses a medial axis skeleton for sampling and coordination guidance. In environments with open spaces, it overly constrains the problem by sampling around the skeleton, leading to a low success rate. While skeleton guidance is ineffective in these scenarios, there is still a benefit to workspace guidance during planning. Similar to CIPHER, sRRT and ARC are hybrid methods that plan decoupled paths for robots and search the larger composite space only when coordination is needed. These methods, however, move into the composite space reactively when inter-robot conflicts occur. By proactively using workspace guidance to avoid conflicts through our decomposition refinement, we see a significant improvement in planning time, even for large groups of robots.
PP-RG-RRT performs similarly to Decoupled RRT, but fails when a higher-priority robot's path blocks another robot's path. This shows the benefit of CIPHER's conflict resolution strategy.

\subsubsection{Kinodynamic Planners}
\label{sec:kino_results}
\begin{figure}
    \centering
    \subfigure[Environment Used]{
        \includegraphics[height=0.25\linewidth]{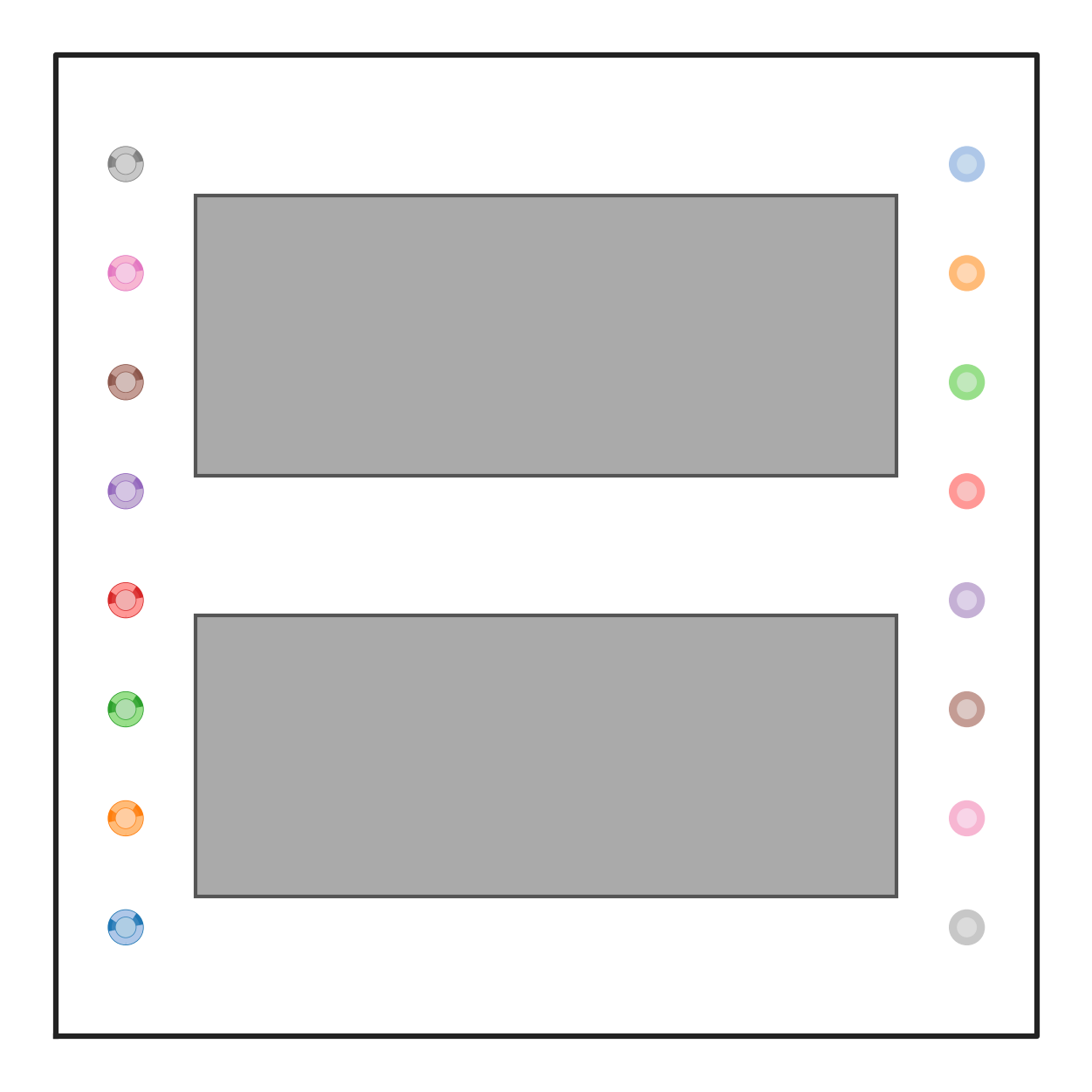}
    }
    \subfigure[Planning Time]{
        \includegraphics[height=0.25\textwidth]{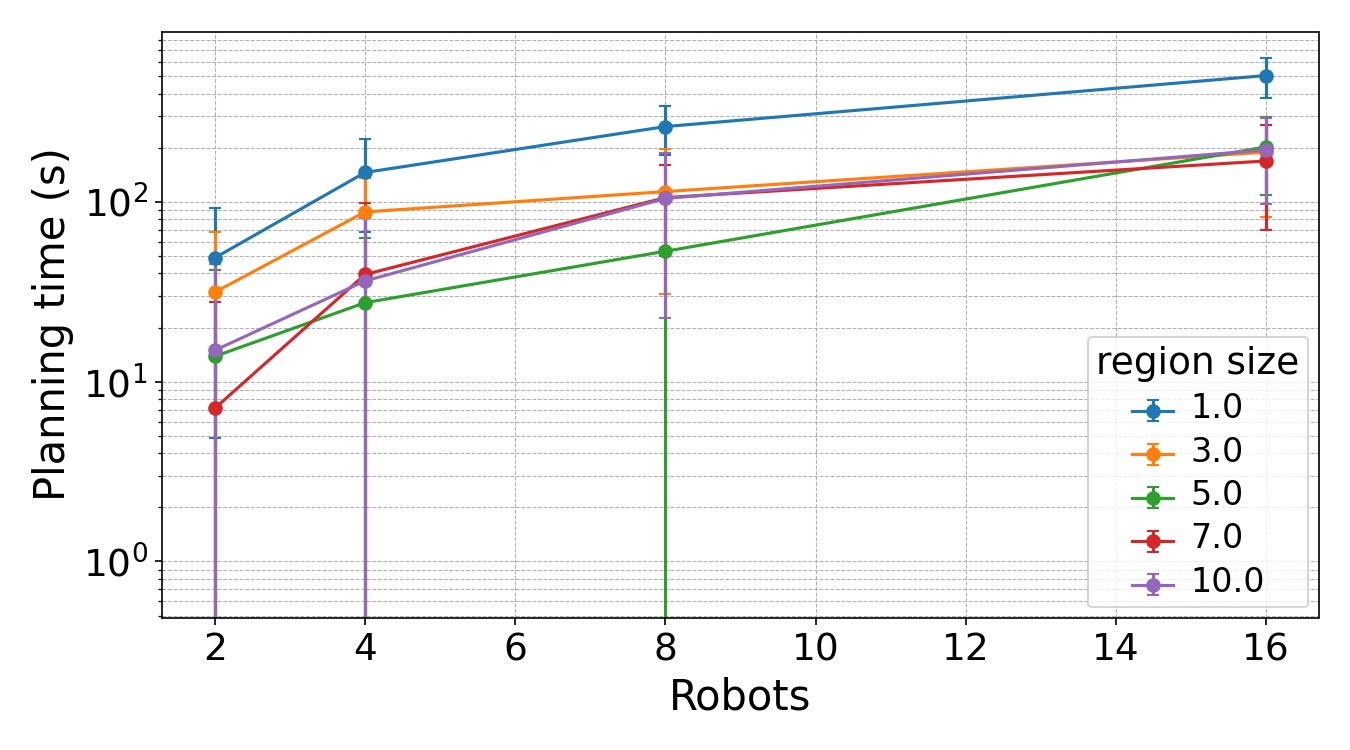}
    }
    \caption{A comparison of region sizes for Kinodynamic CIPHER.}
    \label{fig:mapf-ablation}
    \vspace{-10pt}
\end{figure}

Results for the kinodynamic methods are shown in Fig.~\ref{fig:kinodynamic-results}. CIPHER is again able to solve 100\% of scenarios. Similar trends hold from the geometric results. CIPHER outperforms Coupled RRT, WG-DaSH, and K-ARC in success rate as a result of its ability to leverage and refine the region decomposition to coordinate robots at a high level. Here, PP-RG-RRT, because of its use of guidance to direct planning through these difficult to search kinodynamic planning spaces, achieves the fastest planning times in all environments, up to an order of magnitude faster than Decoupled RRT. However, PP-RG-RRT suffers a low success rate with larger numbers of robots. This highlights a tradeoff between planning time and success rate. While CIPHER is able to solve every scenario due to the time it spends on conflict resolution, PP-RG-RRT saves time by bypassing this step, sometimes resulting in conflicts it cannot resolve. Additionally, we looked at the effect of region size on planning time, Fig~\ref{fig:mapf-ablation} and we found that when the region size is close to the size of the robot, the guided planner struggles to find paths through it.
This potentially means that our adaptive refinement can hinder the ability to solve kinodynamic problems if the minimum region size is close to the size of the robot.
However, if no refinement is performed, the regions may not be sufficient to coordinate higher numbers of robots.

\subsubsection{CIPHER Analysis}
We further analyze the performance of the kinodynamic variant of CIPHER in an environment intended to induce congestion by forcing robots to swap places diagonally across the environment. Fig.~\ref{fig:mapf-ablation} shows a comparison of the performance of CIPHER using different initial region sizes as a multiple of robot size. Region sizes closer to the size of the robot lead to overconstraining the problem when searching for low-level motion paths, especially in kinodynamic problems as discussed in Section~\ref{sec:kino_results}. Larger region sizes (e.g. 10) increase the difficulty of the MAPF step since robots take up a larger area of the workspace by occupying a cell, leading to increased planning times. In our experiments, we use a region size of 5 times the size of the robot.

\section{Conclusions and Future Work}
We present resolution guidance, a multi-robot motion planning approach that refines a region decomposition to coordinate robots motion through the workspace. We propose two implementations of region guidance, PP-RG-RRT and CIPHER. PP-RG-RRT naively explores resolution guidance without conflict resolution. CIPHER iteratively refines the region decomposition representation as a form of conflict resolution. 
Resolution guidance allows our method to reduce the need to plan in the composite space of multiple robots that come into proximity with each other. Our planning scheme achieves improved planning times in both geometric and kinodynamic planning problems relative to other state-of-the-art geometric and kinodynamic planning approaches.

Considering future work, we plan to explore additional approaches to provide coordination between robots at a high-level to further reduce the need for the planner to enter the computationally expensive composite space.
We will also explore different forms of workspace representations including adaptive decompositions that would provide better representations of narrow passages. While our region decomposition does not directly account for obstacles, it may be possible to reduce congestion by using a representation that more explicitly allows the high-level search to route robots around obstacles without constraining them in open environments like a skeleton representation.

%
%
%
\bibliographystyle{splncs04}
\bibliography{references}

\end{document}